\documentclass{jmlr} % test grayscale version
\usepackage{enumitem}
\usepackage[font=small]{caption}

\jmlrvolume{19} % TODO(rliaw): Fix
\jmlryear{2018}
\jmlrworkshop{ICML 2018 AutoML Workshop}

\makeatletter
\def\blfootnote{\xdef\@thefnmark{}\@footnotetext}
\makeatother

\newcommand{\Tune}[0]{Tune}
\title{\Tune{}: A Research Platform for Distributed Model Selection and Training}

\author{\Name{Richard Liaw*} \Email{rliaw@berkeley.edu}
  \AND
  \Name{Eric Liang*} \Email{ericliang@berkeley.edu}
  \AND
  \Name{Robert Nishihara} \Email{rkn@berkeley.edu}
  \AND
  \Name{Philipp Moritz} \Email{pcm@berkeley.edu}
  \AND
  \Name{Joseph E. Gonzalez} \Email{jegonzal@eecs.berkeley.edu}
  \AND
  \Name{Ion Stoica} \Email{istoica@cs.berkeley.edu}
 }

\newif\ifcomments

\commentstrue
\definecolor{purple}{RGB}{128,0,128}
\definecolor{indigo}{RGB}{75,0,130}

\ifcomments
\newcommand{\ion}[1]{\textbf{\textcolor{blue}{Ion: #1}}}
\newcommand{\rkn}[1]{\textbf{\textcolor{purple}{Robert: #1}}}
\newcommand{\ekl}[1]{\textbf{\textcolor{purple}{Eric: #1}}}
\newcommand{\rliaw}[1]{\textbf{\textcolor{magenta}{Richard: #1}}}
\newcommand{\pcm}[1]{\textbf{\textcolor{purple}{Philipp: #1}}}
\newcommand{\royf}[1]{\textbf{\textcolor{red}{Roy: #1}}}
\else
\newcommand{\ion}[1]{\textbf{\textcolor{blue}{}}}
\newcommand{\rkn}[1]{\textbf{\textcolor{purple}{}}}
\newcommand{\ekl}[1]{\textbf{\textcolor{purple}{}}}
\newcommand{\rliaw}[1]{\textbf{\textcolor{magenta}{}}}
\newcommand{\pcm}[1]{\textbf{\textcolor{purple}{}}}
\newcommand{\royf}[1]{\textbf{\textcolor{red}{}}}
\fi

\usepackage{listings}

\usepackage{color}

\definecolor{gray}{rgb}{0.4,0.4,0.4}
\definecolor{dark_green}{rgb}{0.1,0.5,0.1}
\definecolor{dark_blue}{rgb}{0,0,0.7}
\begin{document}

\maketitle

\vspace{-.2cm}

\begin{abstract}

Modern machine learning algorithms are increasingly computationally demanding, requiring specialized hardware and distributed computation to achieve high performance in a reasonable time frame. Many hyperparameter search algorithms have been proposed for improving the efficiency of model selection, however their adaptation to the distributed compute environment is often ad-hoc.
We propose Tune, a unified framework for model selection and training that provides a narrow-waist interface between training scripts and search algorithms. We show that this interface meets the requirements for a broad range of hyperparameter search algorithms, allows straightforward scaling of search to large clusters, and simplifies algorithm implementation.
We demonstrate the implementation of several state-of-the-art hyperparameter search algorithms in Tune. Tune is available at http://ray.readthedocs.io/en/latest/tune.html.

%Today, individuals and organizations largely write their own ad-hoc utilities for hyperparameter search and model selection, but these are often tied to specific search algorithms or deep learning frameworks, lack the flexibility needed to express important optimizations (e.g., early stopping), and don’t scale beyond a single machine.

\end{abstract}

%!TEX root = main.tex
\section{Introduction}
\label{sec:intro}

Machine learning pipelines are growing in complexity and cost. In particular, the model selection stage, which includes model training and hyperparameter tuning, can take the majority of a machine learning practitioner's time and consume vast amounts of computational resources.
Take for example a researcher aiming to train ResNet-101, a convolutional neural-network model with millions of parameters.
Training this model can take around 24 hours on a single GPU, and performing model selection sequentially will take weeks to complete.
Naturally, one would be inclined to train the model on a cluster in a distributed fashion (\cite{goyal2017accurate}) and utilize many machines to perform model selection in parallel.

To this end, the research community has developed numerous techniques for accelerating model selection including those that are sequential (\cite{snoek2012practical}), parallel (\cite{li2016hyperband}), and both (\cite{jaderberg2017population}).
However, each technique is often implemented on its own, tied to a particular framework, is closed source, or perhaps not even reproducible without significant computational resources (\cite{zoph2016neural}).
Further, often times these techniques require significant investment in software infrastructure for the execution of experiments.

\begin{figure}[t]
  \captionsetup{font=scriptsize}
  \centering
  \includegraphics[width=0.9\textwidth]{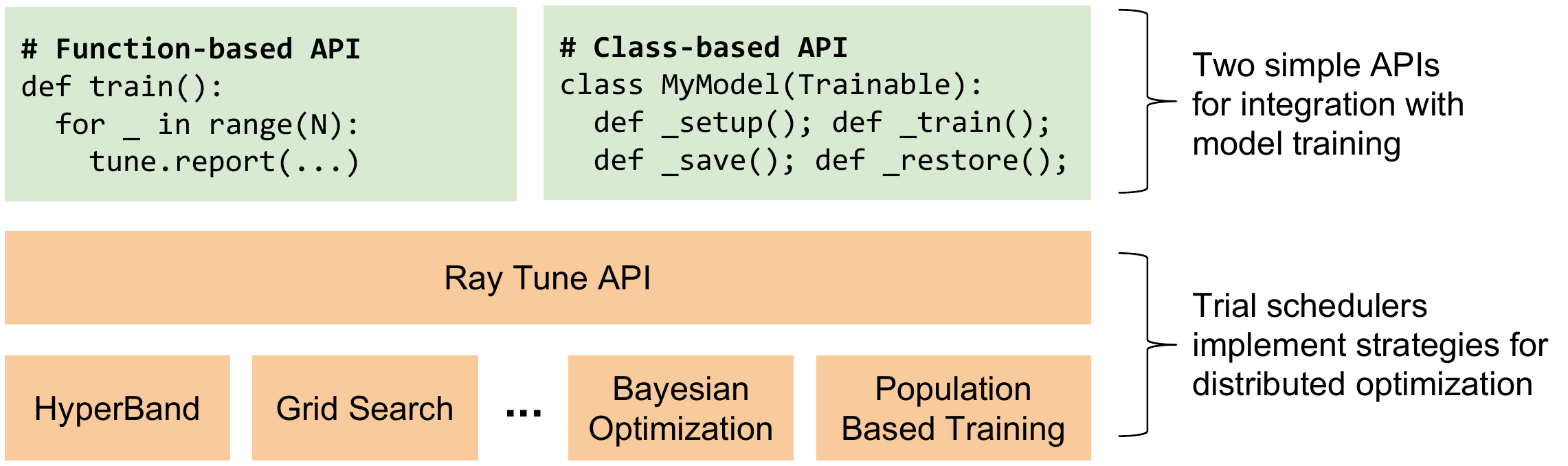}
  \caption[fontsize=11]{Tune provides narrow-waist interfaces that training scripts can implement
  with a few lines of code changes. Once done, this enables the use of Tune for
  experiment management, result visualization, and a choice of trial scheduling strategies.
  This narrow interface also enables \textit{AutoML researchers} to easily swap out different
  search algorithms for comparison, or release distributed implementations of new algorithms
  without needing to worry about distributed scaffolding.
  }
  \label{fig:tune_api}
\end{figure}

The contributions of this paper are as follows:
\begin{enumerate}[noitemsep]
\item We introduce Tune, an open source framework for distributed model selection.
\item We show how Tune's APIs enable the easy reproduction and integration of a wide variety of state-of-the-art hyperparameter search algorithms.
\end{enumerate}

\section{Related work}
\label{sec:related}
There are multiple open source systems for model selection.

HyperOpt, Spearmint, and HPOLib (\cite{snoek2012practical, eggensperger2013towards}) are distributed model selection tools that manage both the search and evaluation of the model, implementing search techniques such as random search and tree of parzen estimators (TPE).
However, both frameworks are tightly coupled to the search algorithm structure and requires manual management of computational resources across a cluster. Further, systems such as Spearmint, HyperOpt, and TuPAQ (MLBase) (\cite{sparks2015automating}) treat a full trial execution as an atomic unit, which does not allow for intermediate control of trial execution. This inhibits efficient usage of cluster resources and also does not provide the expressivity needed to support algorithms such as HyperBand.

% Another popular alternative for distributed model selection is Hyperas, which provides distributed hyperparameter tuning functionality by running using HyperOpt on top of Spark as an execution engine.

Google Vizier (\cite{golovin2017google}) is a Google-internal service that provides model selection capabilities. Similar to Tune, Vizier provides parallel evaluation of trials, hosts many state-of-the-art optimization algorithms, provides functionality for performance analysis. However, it is first and foremost a service and is tied to closed-source infrastructure.

Mistique (\cite{mistique}) is also a system that addresses model selection. However, rather than focusing on the execution of the
selection process, Mistique focuses on model debugging, emphasizing iterative procedures and memory footprint minimization.

Finally, Auto-SKLearn (\cite{feurer2015efficient}) and Auto-WEKA (\cite{thornton2013auto}) are systems for automating model selection, integrating meta learning and ensembling into a single system. The focus of these  systems is at the execution layer rather than the algorithmic level. This implies that in principle, it would be possible to implement Auto-WEKA and Auto-SKLearn on top of Tune, providing distributed execution of these AutoML components. Further, both Auto-SKLearn and Auto-WEKA are tied to Scikit-Learn and WEKA respectively as the only machine learning frameworks supported.

\section{Requirements for API generality}
\label{sec:requirements}

We refer to a \textit{trial} as a single training run with a fixed initial hyperparameter configuration. An \textit{experiment} (similar to a "Study" in Vizier), is a collection of trials supervised by Tune using one of its trial scheduling algorithms (which implement model selection).

A platform for model search and training blends both sequential and parallel computation. During search, many trials are evaluated in parallel. Hyperparameter search algorithms examine trial results in sequence and make decisions that affect the parallel computation. In order to support both a broad range of training workloads and selection algorithms, a framework for model search needs to meet the following requirements:

\begin{itemize}[noitemsep]
\item Ability to handle irregular computations: Trials often vary in length and resource usage.
\item Ability to handle resource requirements of arbitrary user code and third-party libraries. This includes parallelism and the use of hardware resources such as GPUs.
\item Ability to make search / scheduling decisions based on intermediate trial results. For example, genetic algorithms commonly clone or mutate model parameters in the middle of training. Algorithms that perform early stopping also use intermediate results to make stopping decisions.
\end{itemize}

For a good user experience, the following features are also necessary:
\begin{itemize}[noitemsep]
\item The monitoring and visualization of trial progress and outcomes.
\item Simple integration and specification of the experiment to execute.
\end{itemize}

To meet these requirements, we propose the Tune user-facing and scheduling APIs (Section \ref{sec:tune_api}) and implement it on the Ray distributed computing framework (\cite{moritz2017ray}). The Ray framework provides the underlying distributed execution and resource management. Its flexible task and actor abstractions allow Tune to schedule irregular computations and make decisions based on intermediate results.

\section{Tune API}
\label{sec:tune_api}

Tune provides two development interfaces: a \textit{user API} for users seeking to train models and a \textit{scheduling API} for researchers interested in improving the model search process itself. As a consequence of this division, users of Tune have a choice of many search algorithms. Symmetrically, the scheduler API enables researchers to easily target a diverse range of workloads and provides a mechanism for making their algorithms available to users.

\subsection{User API}

%In our experience, the majority of model search users start with simple techniques such as grid search for their ease of interpretability, quickly scale up their search to use multiple machines, and only lastly do they experiment with scheduling algorithms for improving search efficiency. Hence, Tune seeks to provide significant ease of use even for the basic use cases.

%\textbf{Controlling model training}

Model training scripts are commonly implemented as a loop over a model improvement step, with results periodically logged to the console (e.g., every training epoch). To support the full range of model search algorithms, Tune requires access to intermediate training results, the ability to snapshot training state, and also the ability to alter hyperparameters in the middle of training. These requirements can be met with minimal modifications to existing user code via a \textit{cooperative} control model.

\begin{figure}[t]
  \captionsetup{font=scriptsize}
  \centering
  \begin{subfigure}[Function-based API]{
      \label{fig:train_func}
      \includegraphics[width=7cm]{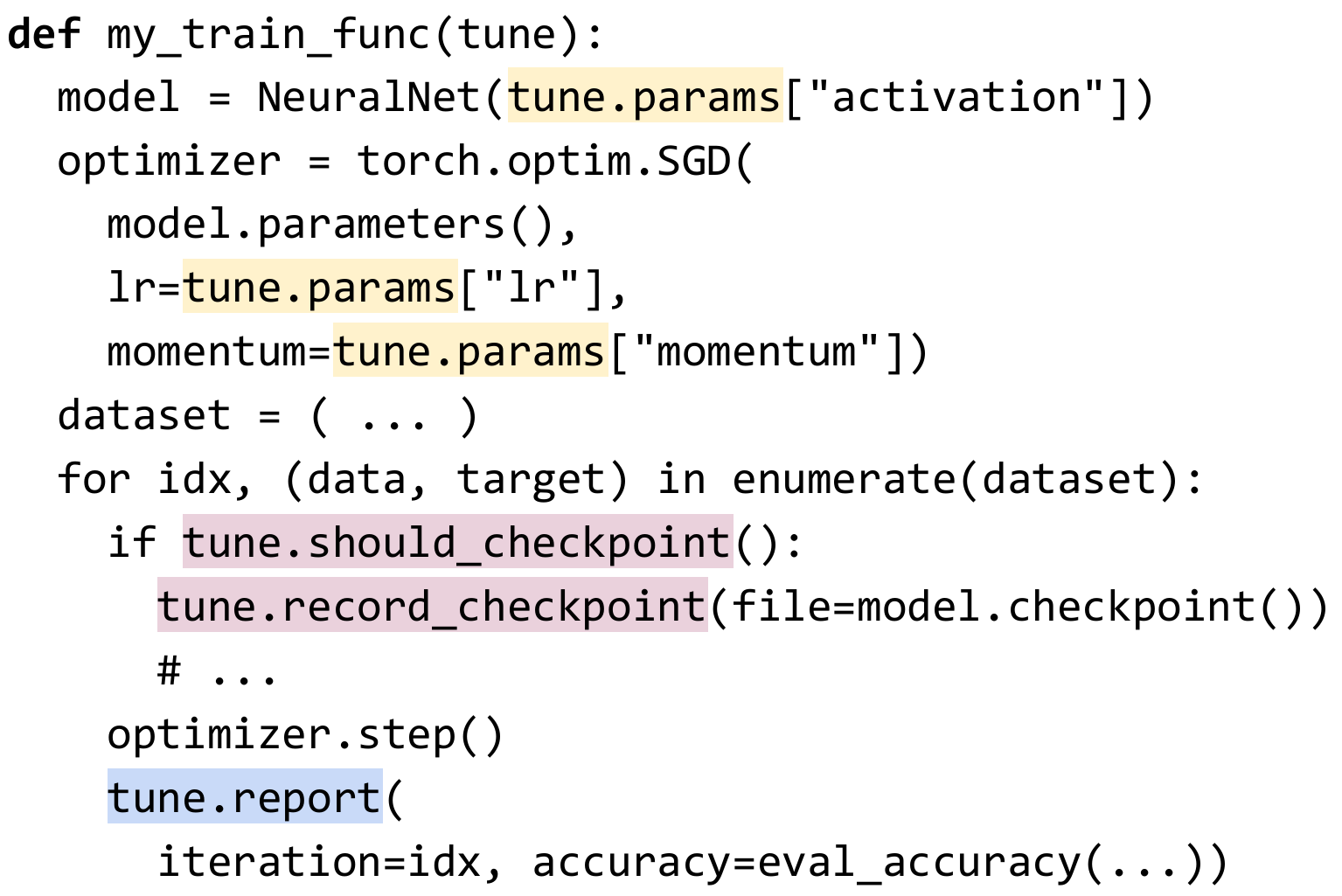}
  }
  \end{subfigure}
  \begin{subfigure}[Class-based API]{
    \label{fig:train_class}
    \includegraphics[width=7cm]{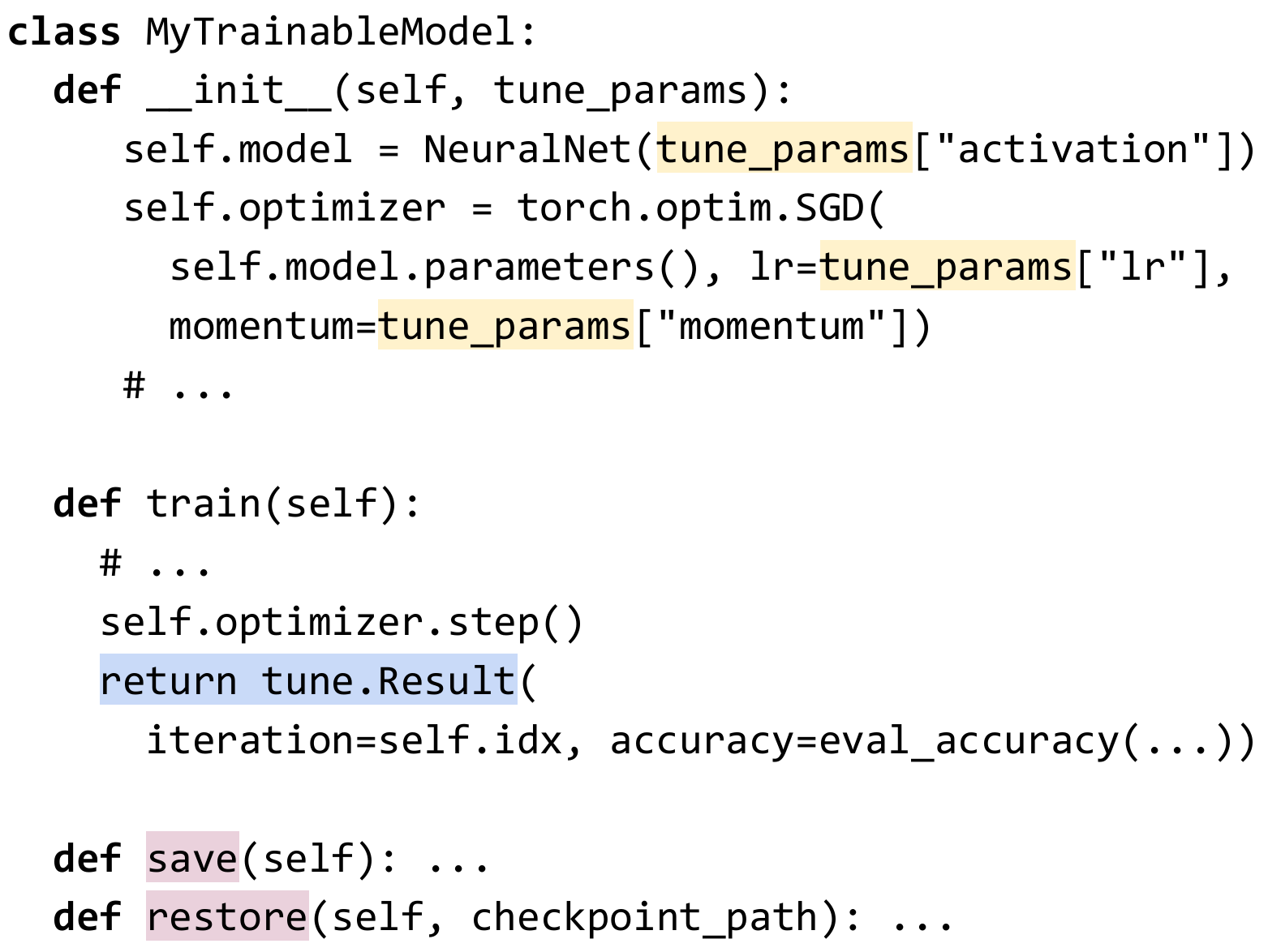}
  }
  \end{subfigure}

    \caption{Tune offers both a function-based \textit{cooperative} control API and
    a class-based API that allows for direct control over trial execution.
    Either can be adopted by the user to enable control over model training via
    Tune's trial schedulers.
    }

\end{figure}

Consider a typical model training script as shown in \ref{fig:train_func}. A handle to Tune is passed into the function. To integrate with Tune, the model and optimizer hyperparameters are pulled from the \textbf{tune.params} map, checkpoints are created when \textbf{tune.should\_checkpoint()} returns positive, and the saved checkpoint file is passed to \textbf{tune.record\_checkpoint()}. Intermediate results are reported via \textbf{tune.report()}. These cooperative calls enable Tune scheduling algorithms to monitor the training progress of each trial (via the reported metrics), save and clone promising parameters (via checkpoint and restore), and alter hyperparameters at runtime (by restoring a checkpoint with a changed hyperparameter map). Critically, these calls require minimal changes to existing user code.

Tune can also \textit{directly control} trial execution if the user extends the trainable model class (Figure \ref{fig:train_class}). Here, training steps, checkpointing, and restore are implemented as class methods which Tune schedulers call to incrementally train models. This mode of execution has some debuggability advantages over cooperative control; we offer both to users. Internally, Tune inserts adapters over the cooperative interface to provide a facade of direct control to trial schedulers.

\subsection{Scheduler API}
\label{sec:alg_api}

Given the ability to create trials and control their execution, the next question is how trials should be scheduled.
Tune's \textit{trial schedulers} operate over a set of possible trials to run, prioritizing trial execution given available cluster resources. In addition, they can add to the list of trials to execute (e.g., based on suggestions from HyperOpt).

The simplest trial scheduler executes trials sequentially, running each until a stopping condition is met. Trials are launched in parallel when sufficient resources are available in the cluster.
However, this is not all trial schedulers can do. They can:

\begin{enumerate}[noitemsep]

\item Early stop a trial that is not performing well based on its intermediate results.

\item Adjust the annealing of hyperparameters such as learning rate.

\item Clone the parameters of a promising trial and launch additional trials that explore the nearby hyperparameter space.

\item Query a shared database of trial results to choose promising hyperparameters.

\item Prioritize resource allocation between a large number of trials, more than can run concurrently given available resources.

\end{enumerate}

The primary interface for a trial scheduler is as follows:

\begin{lstlisting}
class TrialScheduler:
    def on_result(self, trial, result): ...
    def choose_trial_to_run(self): ...
\end{lstlisting}

The interface is event based. When Tune has resources available, it calls \textbf{scheduler.choose\_trial\_to\_run()} to get a trial to launch on the cluster. As results for the trial become available, the \textbf{scheduler.on\_result()} callback is invoked and the scheduler returns an flag indicating whether to continue, checkpoint, stop, or restart a trial with an updated hyperparameter configuration. This interface is sufficient to provide a broad range of hyperparameter tuning algorithms including Median Stopping Rule (\cite{golovin2017google}), Bayesian Optimization approaches (\cite{bergstra2013making}), HyperBand (\cite{li2016hyperband}), and Population-based Training (\cite{jaderberg2017population}).

We note that Tune keeps the metadata for active trials in memory and relies on checkpoints for fault tolerance. This drastically simplifies the design of trial schedulers and is not a limitation in practice. Trial scheduler implementations are free to leverage external storage if necessary.

\subsection{Putting it together}

To launch an experiment, the user must specify their model training function or class (Figures \ref{fig:tune_api} and \ref{fig:train_class}), an initial set of trials, and a trial scheduler. The following is a minimal example:

\begin{lstlisting}
def my_func(): ...
tune.run_experiments(my_func, {
    "lr": tune.grid_search([0.01, 0.001, 0.0001]),
    "activation": tune.grid_search(["relu", "tanh"]),
}, scheduler=HyperBand)
\end{lstlisting}

Here, we use Tune's built-in DSL to specify a small $3\times2$ grid search over two hyperparameters. These serve as the initial set of trials input to the scheduler. Tune's parameter DSL offers features similar to those provided by HyperOpt (\cite{bergstra2013making}). Alternatively, users can generate the list of initial trial configurations with the mechanism of their choice. Once an experiment is launched, the progress of trials is periodically reported in the console and can also be viewed through integrations such as TensorBoard.

\subsubsection{Scaling computation}

Each trial in Tune runs in its own Python process, and can be allocated given number of CPU and GPU resources through Ray. Individual trials can themselves leverage distributed computation by launching further subprocesses using Ray APIs. These child processes can coordinate with each other using Ray to e.g., perform SGD, or use collective communication primitives provided by libraries such as \texttt{torch.distributed} and Nvidia NCCL.

\subsubsection{Data input}

Since each trial in Tune runs as a Ray task or actor, they can use Ray APIs to handle data ingest. For example, weights can be broadcast to all workers using \textbf{ray.put(obj)} to the Ray object store, and retrieved via \textbf{ray.get(obj\_id)} during trial initialization.

\section{Implementation}
\label{sec:evaluation}

\begin{table}
\captionsetup{font=scriptsize}
\centering
\begin{tabular}{|c|c|}
\hline
\bfseries Algorithm & \bfseries Lines of code\\
\hline
FIFO (trivial scheduler) & 10 \\
Asynchronous HyperBand (\cite{li2018massively})& 78 \\
HyperBand (\cite{li2016hyperband}) & 215 \\
Median Stopping Rule & 68 \\
HyperOpt (\cite{bergstra2013making}) & 137 \\
Population-Based Training (\cite{jaderberg2017population})& 169 \\
\hline
\end{tabular}
\caption{Model selection algorithms implemented (or integrated) in Tune. Two versions of HyperBand are implemented: the original formulation and the asynchronous variation which is simpler to implement in the distributed setting.}
\label{table:tune_loc}
\end{table}

We list in Table \ref{table:tune_loc} currently implemented algorithms in Tune. Line counts include lines used for logging and debugging functionality.
% \subsection{Execution Engine}
We implemented Tune using the Ray (\cite{moritz2017ray}) framework, which as noted earlier provides the actor abstraction used to run trials in Tune. In contrast to popular distributed frameworks such as Spark (\cite{spark}), or MPI (\cite{mpi}), Ray offers a more flexible programming model. This flexibility enables Tune's trial schedulers to centrally control the many types of stateful distributed computations created by hyperparameter optimization algorithms.

The Ray framework is also uniquely suited for executing nested computations (i.e., hyperparameter optimization) since it has a \textit{two-level} distributed scheduler. In Ray, task scheduling decisions are typically made on the local machine when possible, only "spilling over" to other machines on the cluster when local resources are exhausted. This avoids any central bottleneck when distributing trial executions that may themselves leverage parallel computations.

%\subsection{Median Stopping Rule}
%
%Pseudocode for Median Stopping Rule scheduler:
%\begin{verbatim}
%def on_result(self, trial, result):
%    if self.is_below_median(result):
%        return Trial.STOP
%    else:
%        self.store_result(trial, result)
%        return Trial.CONTINUE
%\end{verbatim}

%\subsection{Population-based Training}
%\label{sec:pbt}
%Population-based training (\cite{jaderberg2017population}) is a model search algorithm that alters the hyperparameter configuration of a given model during the trial and shares information across other concurrent trials with a parameter transfer scheme. To provide more intuition on how trial schedulers function, below we include pseudocode for the PBT trial scheduler:

%\begin{lstlisting}
%def on_result(self, trial, result):
%    if self.is_top(trial, result):
%        trial_checkpoint[trial] = trial.checkpoint()
%    else:
%        del trial_checkpoint[trial]
%    self.store_result(trial, result)
%    if self.is_bottom(trial, result):
%        trial.restore(mutate(random.choice(trial_checkpoints)))
%    return Trial.CONTINUE
%\end{lstlisting}

%\input{arch.tex}

\section{Conclusion and Future Work}

In this work, we explored the design of a general framework for hyperparameter tuning.
We proposed the Tune API and System which supports extensible distributed hyperparameter search algorithms while also being easy for end-user model developers to incorporate into their model design processes.
We are actively developing new functionality to help not only in the tuning process but also in analyzing and debugging the intermediate results.

% \section*{References}

% \small

% [1] Alexander, J.A.\ \& Mozer, M.C.\ (1995) Template-based algorithms
% for connectionist rule extraction. In G.\ Tesauro, D.S.\ Touretzky and
% T.K.\ Leen (eds.), {\it Advances in Neural Information Processing
%   Systems 7}, pp.\ 609--616. Cambridge, MA: MIT Press.

% [2] Bower, J.M.\ \& Beeman, D.\ (1995) {\it The Book of GENESIS:
%   Exploring Realistic Neural Models with the GEneral NEural SImulation
%   System.}  New York: TELOS/Springer--Verlag.

% [3] Hasselmo, M.E., Schnell, E.\ \& Barkai, E.\ (1995) Dynamics of
% learning and recall at excitatory recurrent synapses and cholinergic
% modulation in rat hippocampal region CA3. {\it Journal of
%   Neuroscience} {\bf 15}(7):5249-5262.

\pagebreak
\bibliography{main}
% \bibliographystyle{jmlr}
% \bibliographystyle{IEEEtranN}
% \pagebreak
% \onecolumn{
% \appendix

% % \input{appendix}
% }

\end{document}